\pdfoutput=1
\documentclass[sigconf, screen, nonacm]{acmart}
\AtBeginDocument{%
  }

\setcopyright{none}
\usepackage{xspace}
\newcommand{\sys}{\textsc{Cortex-FL}\xspace}

\usepackage{pifont}
\newcommand{\stage}[1]{\ding{\numexpr181+#1\relax}}

\begin{document}

\title{Cortex-FL: Agent-Mediated Collaboration for Federated Learning}

\settopmatter{authorsperrow=3}

\author{Rui Sun}
\affiliation{
  \institution{Newcastle University}
  \city{Newcastle upon Tyne}
  \country{UK}
}
\email{rui.sun@newcastle.ac.uk}

\author{Jinfeng Peng}
\affiliation{
  \institution{FLock.io}
  \city{London}
  \country{UK}
}
\email{leo@flock.io}

\author{Zehua Cheng}
\affiliation{
  \institution{University of Oxford}
  \city{Oxford}
  \country{UK}
}
\additionalaffiliation{
  \institution{Korea University}
  \city{Seoul}
  \country{South Korea}
}
\email{zehuac@korea.ac.kr}

\author{Jiahao Sun}
\affiliation{
  \institution{FLock.io}
  \city{London}
  \country{UK}
}
\email{sun@flock.io}

\author{Zhipeng Wang}
\affiliation{
  \institution{The University of Manchester}
  \city{Manchester}
  \country{UK}
}
\email{zhipeng.wang@manchester.ac.uk}

\renewcommand{\shortauthors}{Sun et al.}


\begin{abstract}
Federated learning (FL) enables independent data owners to train models collaboratively without centralizing raw data. However, identifying collaborators, preparing training tasks, and coordinating execution and incentives hinder its practical adoption by organizations without specialized expertise. We present \sys{}, an agent-native FL system that enables users to express learning objectives in natural language and delegate the collaboration required to train a model. Agents translate these objectives into shared tasks, discover collaborators, and negotiate participation agreements, connecting collaboration formation with distributed training and incentive management. The architecture separates agent coordination from local execution authority, allowing owners to delegate coordination while retaining control over their data, computation, and financial commitments. We demonstrate the complete workflow on a four-node deployment, in which a model requested in natural language proceeds through collaboration formation, contract-governed training, verified model delivery, and x402 settlement on a blockchain testnet.
\end{abstract}




\maketitle

\section{Introduction}
\label{sec:introduction}

The emerging Agentic Web envisions autonomous AI agents that coordinate tools, services, and computational resources on behalf of users~\cite{sun2025vision,yang2025agenticweb}. Systems such as Claude Code~\cite{anthropic2025claudecode} and Codex~\cite{openai2025codex} demonstrate natural-language delegation of complex software engineering tasks. At the network level, the Internet of Agents~\cite{chen2025ioa} and the Decentralized Agent Swarm Network~\cite{sun2025vision} explore collaboration among independently operated agents with heterogeneous capabilities. These developments motivate extending delegation to collaborative learning across organizational boundaries, where training relies on data and computation held by multiple owners.

Federated learning (FL) enables such collaboration by allowing independent data owners to train shared models without centralizing raw data~\cite{mcmahan2017fedavg}. Its practical deployment, however, requires coordination beyond the training algorithm. In cross-silo settings, organizations must align learning objectives, configure compatible execution environments, and establish participation terms and incentives~\cite{kairouz2021advances}. Data owners seeking a model may lack established collaborators or the expertise to operate the required infrastructure, which hinders adoption by organizations without dedicated FL expertise. Addressing this barrier requires supporting the formation and management of a federation alongside distributed training.

\begin{figure}[htp]
\centering
\includegraphics[width=\columnwidth]{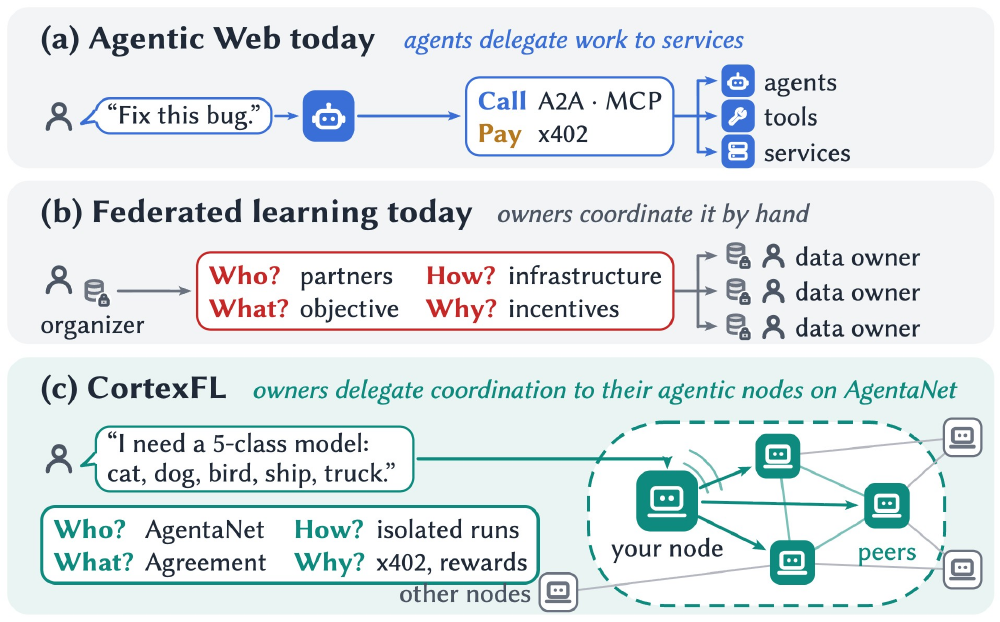}
\caption{From (a) agentic service delegation and (b) manually coordinated FL to (c) agent-mediated FL in \sys{}.}
\label{fig:problem}
\Description{Three panels compare coordination workflows. Panel (a) shows an agent delegating a software task to agents, tools, and services through A2A and MCP, with x402 payments. Panel (b) shows an organizer coordinating data owners through questions about participants, learning objectives, infrastructure, and incentives. Panel (c) shows CortexFL agentic nodes on AgentaNet coordinating a classifier for cats, dogs, birds, ships, and trucks through agreements, isolated training, and incentives.}
\end{figure}

Extending agent delegation to this setting requires connecting general interoperability mechanisms with the FL lifecycle. A2A~\cite{a2a2025} and MCP~\cite{mcp2024} support inter-agent communication and tool integration, while x402~\cite{x402protocol} enables service payments and ERC-8004~\cite{erc8004} specifies registries for agent identity and reputation. FL additionally requires linking participant discovery to shared task specifications, participation agreements, and coordinated execution across training rounds. The challenge is therefore to integrate collaboration formation, distributed training, and incentive management while preserving each owner's authority over local data, computation, and payments. Figure~\ref{fig:problem} contrasts agentic service delegation and manual FL coordination with the integrated workflow of \sys{}.

We present \sys{}, an agent-native decentralized FL system that enables independent data owners to initiate and manage cross-silo collaborations through natural-language instructions (Figure~\ref{fig:problem}c). Built on Cortex\footnote{\url{https://cortex.flock.io}}, a desktop workbench for AI agents, \sys{} represents each data owner as an independently controlled \emph{agentic node} (AN). Agents discover candidate collaborators, prepare executable learning tasks, and coordinate participant admission and distributed training. The local host integrates authenticated coordination, package integrity verification, and sandboxed execution, reducing manual configuration. Raw training data remains local, while publication, participation, execution, and payments require owner authorization. The system integrates x402 service payments with FL~Alliance training commitments and rewards, keeping service transactions distinct from participation incentives. Our demonstration follows the complete workflow from a natural-language objective to collaborative training, model inspection, and service settlement through the desktop interface.

\section{Design Requirements and Trust Model}
\label{sec:goals}

\noindent\textbf{Design requirements.}
\sys{} targets cross-silo federations whose members independently administer their data and computing resources. Four requirements guide the design. \emph{(R1) Low-overhead deployment} reduces setup effort through reusable node hosts and managed training environments. \emph{(R2) Agent-mediated collaboration} connects natural-language objectives to task preparation, peer discovery, and participation agreements. \emph{(R3) Owner-controlled execution} binds collaboration to authenticated identities, verified artifacts, and approved execution scope. \emph{(R4) Accountable incentives} establishes explicit commitments and maintains verifiable records of both training rewards and service payments.

\noindent\textbf{Trust model.}
Each owner trusts the local host, authorization interface, and credential store. Peer messages, retrieved artifacts, and agent-visible content are treated as untrusted inputs. The host validates proposed actions against the task specification, authenticated identities, and current authorization. Publication, participation, execution, and payments require explicit approval or an owner-approved policy covering the action, amount, and deadline. Isolated workers access task-scoped data and invoke restricted host operations for external communication. Wallet and service credentials remain under host custody, providing a common authorization boundary. These guarantees cover owner authorization, artifact integrity, and transaction accounting across every agent action and every training workload on the node.

\section{System Overview}
\label{sec:overview}

\begin{figure*}[htp]
\centering
\includegraphics[width=\textwidth]{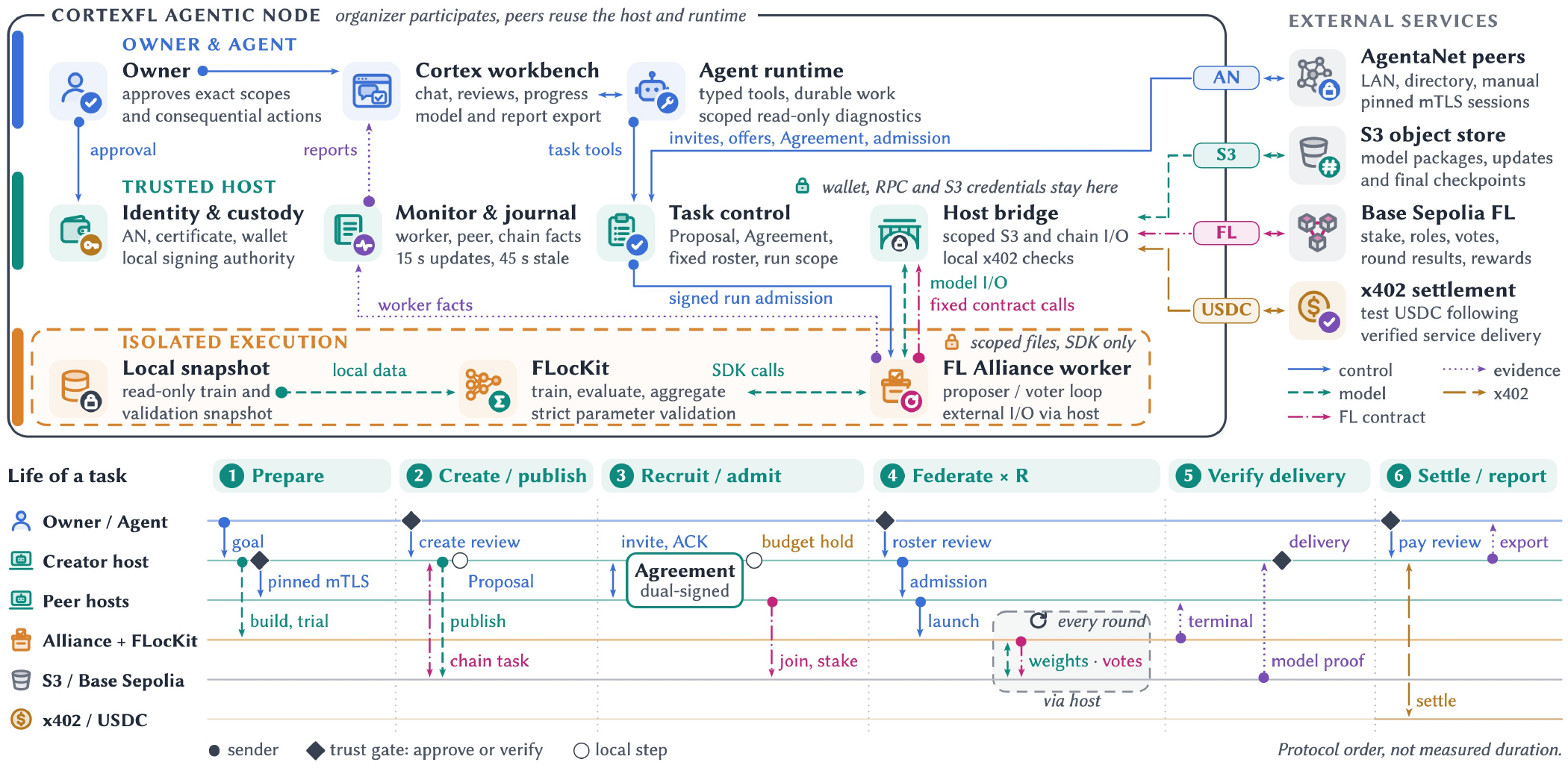}
\caption{Architecture (top) and task lifecycle (bottom) of \sys{}.}
\label{fig:arch}
\Description{The upper panel separates owner and agent interaction, trusted host coordination, and isolated training. The host mediates connections to AgentaNet peers, S3 storage, Base Sepolia contracts, and x402 settlement. The lower panel follows preparation, task publication, participant admission, federated training, delivery verification, and settlement across the owner and agent, creator host, peer hosts, training runtime, storage and chain, and payment service.}
\end{figure*}

As shown in Figure~\ref{fig:arch}, each agentic node comprises an interaction layer, a trusted host, and an isolated training runtime. The Cortex workbench combines natural-language interaction with action review and result inspection. Agents invoke typed tools for task preparation, negotiation, and monitoring. The host maintains task state and executes authorized operations through scoped interfaces. Within the isolated runtime, FL~Alliance executes the round protocol and invokes FLocKit for model computation. The organizer can also participate in training, and every node retains its own identity, local data, and execution records. Figure~\ref{fig:arch} (bottom) follows a task between the organizer (\emph{creator}) and its peers through six stages, \stage{1}--\stage{6}, which structure the rest of this section.

\noindent\textbf{Design highlights.}
\sys{} reuses FL~Alliance and FLocKit for training, x402 for payment, and AgentaNet for peer discovery. Its own contribution is the layer that connects them, which rests on three design choices. First, agents coordinate without holding authority. They plan, negotiate, and report but hold no wallet or service credentials, and every consequential step passes a trust gate (diamonds in Figure~\ref{fig:arch}) that requires owner approval or host verification. Second, a signed artifact chain binds the manifest, the bilateral agreements, and the run admission to the task registered on chain, so each node can verify what it is asked to run. Third, incentives follow two separate tracks. FL contracts govern protocol rewards and penalties, while service fees are reserved before admission and released through x402 only after the buyer has verified delivery.

\subsection{Agent Tools and Owner Authority}
\label{sec:authority}

The agent acts only through typed tools that mirror the task lifecycle, from drafting and trial runs to negotiation and payment. Tool arguments carry opaque task and dataset identifiers, so even a misled agent cannot name a host path, an account, or a transaction field, and local data enters only through a native file dialog operated by the owner. A call that publishes, joins, trains, or pays raises an owner review that only the live application window can answer. Payments require an explicit owner click in every permission mode. For repeated steps, the owner can approve a bounded policy, and no policy can approve or enlarge itself.

\subsection{Agent-Mediated Collaboration}
\label{sec:admission}

\textbf{Prepare and publish (\stage{1}, \stage{2}).}
\sys{} supports different learning tasks through model templates that supply training, aggregation, and evaluation routines, covering image classification, graph node classification, time-series prediction, and LLM fine-tuning. The organizer's agent configures a template according to the user's objective and owner-authorized data. The host builds the model package, binds the task to a versioned data snapshot, and checks data compatibility, such as class coverage and sample counts, before a bounded local trial. Following owner approval, it registers the task and model hash on Base Sepolia and verifies the creation receipt. A signed manifest (the \emph{Proposal} in Figure~\ref{fig:arch}) fixes the package digest, configuration, execution backend, weight format, and training parameters, providing a common reference for verification and execution on every participating node.

\textbf{Recruit and admit (\stage{3}).}
AgentaNet supplies candidate peers through local discovery, a directory, or specified endpoints, with signed identity exchanges and pinned mutual TLS authenticating the connections. Each candidate verifies the package against the approved source, manifest, and registered hash, then prepares its own data snapshot and performs a local trial. These checks establish local readiness under the common specification, while participants' datasets may differ in size and distribution. Signed offers and acceptances bind the parties, task, and execution limits in bilateral FL agreements. After checking eligibility, funding, capacity, and any service budget commitment described in Section~\ref{sec:pay}, the organizer obtains owner approval for a fixed roster and issues signed run admission binding member identities and certificates.

\subsection{Contract-Governed Training}
\label{sec:incentives}

In stage \stage{4}, \sys{} coordinates model updates and participant incentives through FL~Alliance's training protocol, which builds on the stake-based model validation and incentive mechanisms introduced in FLock~\cite{dong2024flock}. Its smart contracts govern staking, role assignment, round transitions, rewards, and slashing. Each host runs the protocol in an isolated FL~Alliance worker within its approved execution scope, which invokes the training, aggregation, and evaluation routines of the agreed package through FLocKit.

In each round, proposer nodes train from the designated checkpoint and submit references to their model updates. Voter nodes validate the retrieved parameters, including tensor names, shapes, data types, finite values, and bounds, before invoking the task's aggregation routine. Each voter evaluates the resulting candidate and the round baseline on the same local validation split and submits its vote through the contract's commit--reveal procedure. The contract-selected checkpoint becomes the next round's reference. Model packages and weights are exchanged through S3 and verified by hash, while raw training samples remain local.

\subsection{Service Payments and Settlement}
\label{sec:pay}

Participation by independent providers may depend on service fees negotiated separately from FL~Alliance's protocol rewards and penalties. FL agreements themselves carry no fee. \sys{} therefore maintains optional service orders linked to the corresponding FL agreements. Signed quotes and counteroffers record the buyer, provider, agreed price, and execution and collection deadlines. Before training, the buyer reserves the agreed amount in its local ledger and sends the provider a signed commitment, which the provider verifies before service admission. The reservation records planned expenditure, while payment authorization still requires delivery verification and owner approval.

In stage \stage{5}, the buyer checks the final checkpoint against the contract-selected reference and verifies delivery using the original order, retained execution evidence, and canonical task result. In stage \stage{6}, following verification and owner approval, it issues an ERC-3009 transfer authorization~\cite{eip3009} for x402~\cite{x402protocol} settlement in test USDC. The host then checks the authorization nonce, recipient, and amount against the canonical transaction receipt before recording settlement, so a payment counts as settled only with on-chain proof. The task report links each service order to its delivery record and settlement receipt, separately from FL stake and rewards.

\section{Implementation}
\label{sec:implementation}

\sys{} is implemented as a dedicated FL module within the Cortex client, built on Electron. The client and the standalone CortexFL Node participant host share the implementation of task coordination, training, and settlement verification, letting the organizer train locally through the same procedures as other participants. The trusted host manages the training environment associated with each approved task. It verifies the model package, configuration, and dependency information before launching an isolated FL~Alliance\footnote{\url{https://github.com/FLock-io/FL-Alliance-Client}} worker using a pinned Python environment managed by \texttt{uv}. The worker invokes the FLocKit\footnote{\url{https://github.com/FLock-io/FLocKit}} SDK to load model parameters and execute the training, aggregation, and evaluation routines supplied by the agreed model package. The training runtime supports MLX on macOS and GPU execution on Linux, with the selected backend and weight format fixed in the task manifest. Worker and model sources ship as bundled packs whose per-file digests are pinned in a manifest, and the approved sources are compiled into the client.

The worker retrieves checkpoints, uploads model updates, and submits contributions and votes through the host bridge shown in Figure~\ref{fig:arch}. The bridge exposes scoped S3 operations and fixed contract calls, through which the host performs artifact transfers, content verification, and authorized transaction signing. A peer can read an object only with the task's signed run admission and the matching mTLS certificate, so knowing a digest grants nothing. Wallet, RPC, and S3 credentials stay with the host, while model computation runs on the task's data snapshot in isolation.

The trusted host records these operations in persistent journals that associate signed agreements, execution attempts, model artifacts, and payment authorizations with their originating tasks. Following interruptions, the host uses the retained records to verify prior receipts and reconcile unresolved operations. Resuming a stopped or failed run requires separate owner authorization. The same records support task reports that associate metrics with the task specification, local data snapshot, execution attempt, and evaluation split. These reports accompany the verified checkpoint and a manifest of file hashes on export.

\section{Demonstration}
\label{sec:demo}

\noindent\textbf{Setup.}
We demonstrate \sys{} with a five-class image classification task on CIFAR-10~\cite{krizhevsky2009learning}, covering cats, dogs, birds, ships, and trucks. The organizer's training data contains only cats and dogs, while peers hold examples of the remaining classes, motivating collaboration across independently held datasets. The deployment comprises a MacBook Pro as organizer and participant and three Mac minis running standalone participant hosts. The devices share a local network, with Internet access to the language-model provider, S3, and the public Base Sepolia testnet.

\begin{figure}[htp]
\centering
\includegraphics[width=\columnwidth]{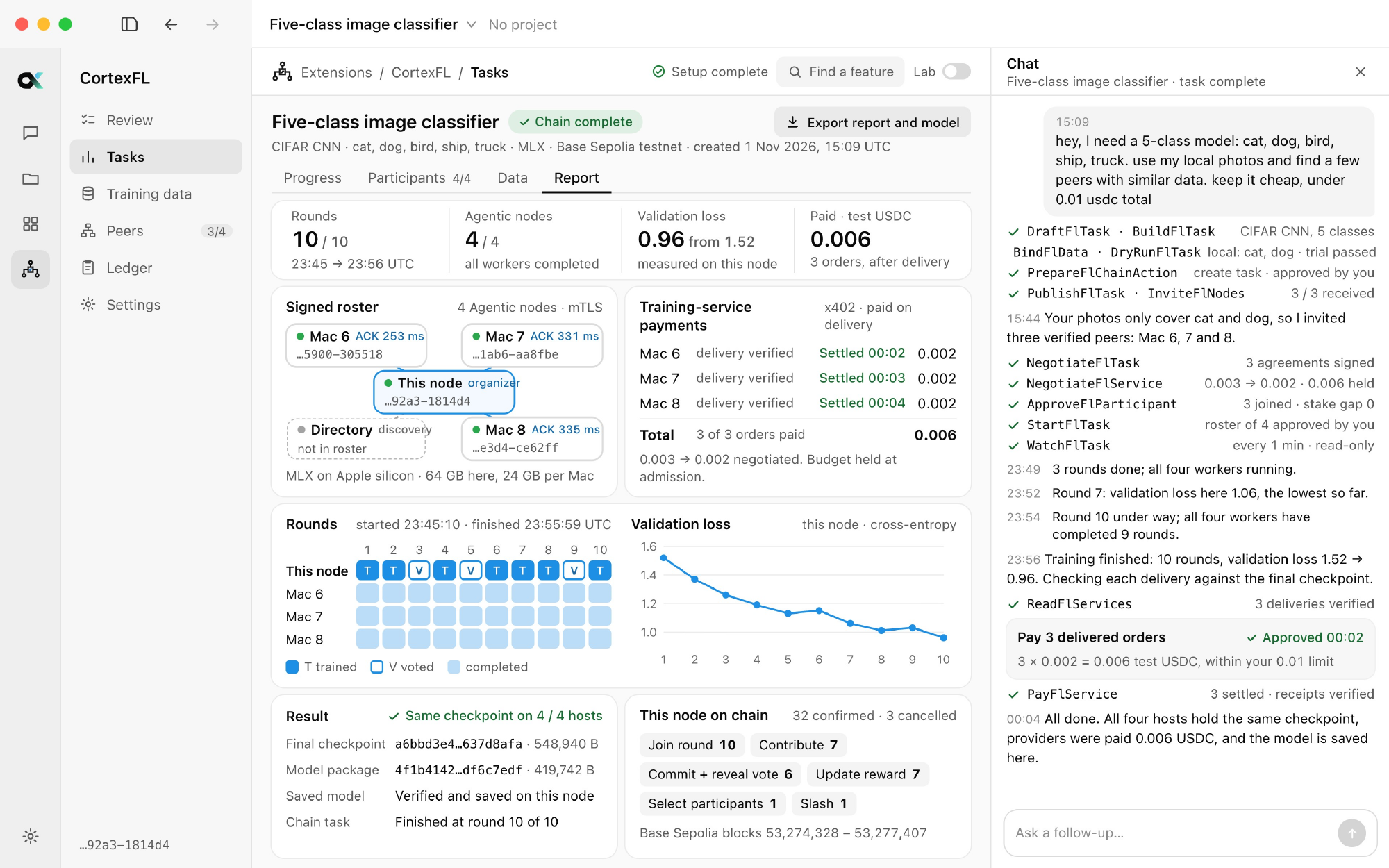}
\caption{The \sys{} workspace, with the task report (roster, round timeline, validation loss, payments, and on-chain activity) on the left and the agent chat on the right.}
\label{fig:screenshot}
\Description{The desktop interface combines an agent chat panel with a task workspace displaying the participant roster, training progress, validation metrics, service payments, contract activity, and model outputs.}
\end{figure}

\noindent\textbf{Walk-through.}
Through the interface in Figure~\ref{fig:screenshot}, the owner specifies the target classes, training rounds, and service budget in natural language. The agent prepares the task and invokes a local trial, whose results the owner reviews before authorizing publication. It then discovers peers and coordinates participation agreements and service orders. To exercise the integrity check, an agent modifies the model package before delivery, and the receiving node rejects it at verification. The owner inspects the agreed execution scope and budget reservations before approving the roster. During training, the interface displays participant activity, round progress, validation metrics, and contract events. The selected template uses equal-weight parameter averaging, and each voter approves a candidate with lower cross-entropy than the round baseline on its local validation split. At completion, the owner inspects the verified checkpoint and its task and round records, then reviews the delivery evidence associated with each service order and authorize payment after verification. The interface presents service settlement receipts separately from FL~Alliance incentive records, and the verified checkpoint and associated task report can be exported from the same workspace.

\section{Related Work}
\label{sec:related}

Flower~\cite{beutel2020flower}, FedML~\cite{he2020fedml}, and FATE~\cite{liu2021fate} provide infrastructure for distributed training. Swarm Learning~\cite{warnat2021swarm} supports blockchain-based coordination, while FLock~\cite{dong2024flock} develops stake-based model validation and incentives. The Internet of Agents~\cite{chen2025ioa} supports heterogeneous agent collaboration through dynamic team formation, and AgentaNet~\cite{sun2025vision} proposes a decentralized architecture connecting agentic nodes for discovery, negotiation, and incentivized service exchange. Building on FLock's mechanisms through FL~Alliance and on AgentaNet, \sys{} carries a learning objective from task preparation and participant recruitment to authorized training, verified model delivery, and service settlement.

\section{Ethical Use of Data}
\label{sec:ethics}

The demonstration uses public CIFAR-10 images. Training data remains on participating devices, and owners authorize its use. Financial transactions use test USDC on the Base Sepolia testnet.

\section{Conclusion}
\label{sec:conclusion}

We presented \sys{}, an agent-native FL system that translates natural-language learning objectives into shared tasks, participation agreements, and coordinated training. By separating agent coordination from local execution authority, the system supports delegation throughout the learning lifecycle while preserving owner control over data, computation, and payments, providing a concrete realization of owner-controlled, agent-mediated collaborative learning within the emerging Agentic Web.

\bibliographystyle{ACM-Reference-Format}
\balance
\bibliography{my_revised}

\end{document}